\documentclass[letterpaper, 10 pt, conference]{ieeeconf}

\IEEEoverridecommandlockouts
\usepackage{booktabs}
\usepackage{makecell}
\usepackage{threeparttable}
\usepackage{graphicx}
\usepackage{amsmath}
\usepackage{amssymb}
\usepackage[table]{xcolor}
\makeatletter
\let\NAT@parse\undefined
\makeatother
\usepackage[numbers,sort&compress]{natbib}
\usepackage{xspace}
\usepackage{subcaption}
\usepackage{pifont}
\usepackage[font=small]{caption}
\usepackage{float}
\usepackage{hyperref}

\title{\LARGE \bf
GGraph: A GPU-Accelerated Library for
Graph-Based Robot Deployment and Learning
}

\author{Yash Rampuria$^{1}$, Rohan Garg$^{1}$, Daniele Gammelli$^{1}$ and Marco Pavone$^{1}$\\
$^{1}$Stanford University, CA
\thanks{Emails: {\tt\small yashr@stanford.edu, rohang73@stanford.edu, gammelli@stanford.edu, pavone@stanford.edu}}%
}

\definecolor{ggcolor}{RGB}{36,44,11}
\newcommand{\sysname}{\textcolor{ggcolor}{\textsc{GGraph}}\xspace}
\newcommand{\cmark}{\textcolor{green!70!black}{\ding{51}}}
\newcommand{\xmark}{\textcolor{red}{\ding{55}}}
\newcommand{\otick}{\textcolor{orange}{\ding{51}}}
\hypersetup{colorlinks=true,linkcolor=blue,citecolor=blue,urlcolor=blue}

\begin{document}
\maketitle
\thispagestyle{empty}
\pagestyle{empty}


\begin{abstract}
Autonomous mobile robots require spatial representations that capture geometry
and semantics while staying sparse enough for planning in large, unknown
environments. Graphs provide a natural representation for this purpose and underpin both classical planning methods and a growing class of reinforcement-learning (RL) policies. However, RL
training needs hundreds to thousands of parallel environments, each constructing
and updating its own graph. Existing graph libraries are typically designed for either single-robot
deployment or fixed graphs, and CPU-based graph construction limits scalable training to only a
few dozen environments. We introduce \sysname, an open-source GPU-accelerated
graph library for both robot deployment and large-scale learning. \sysname supports real robots through ROS~2 interfaces and vectorized GPU-based policy training, with built-in support for tasks including 3-D navigation, exploration, and loop-closure integration. For single-robot operation, \sysname updates graphs in milliseconds. At scale, it enables graph-based exploration policies to train across hundreds of parallel robots, reducing wall-clock training time by one order of magnitude. Across simulation and real-world experiments, these results establish \sysname as a unified graph infrastructure for both large-scale robot learning and deployment.
\end{abstract}


\section{Introduction}

Progress in mobile robotics has been driven by a small set of \emph{design tools} that became shared infrastructure, including occupancy grids for planning \citep{elfes1989using}, GPU elevation mapping for locomotion \citep{miki2022elevation}, OpenCV for perception \citep{opencv_library}, PyTorch for learning \citep{DBLP:journals/corr/abs-1912-01703}, and massively parallel simulators for RL \citep{makoviychuk2021isaac}. Each is valuable because one implementation serves both deployment on a robot and training in simulation, so improvements on either side transfer.

Graphs have long been used for robot planning. A navigation graph stores sparse collision-free nodes connected by traversable edges, and nodes can carry frontiers, traversability, semantics or learned scores. Such graphs underpin classical exploration planners \citep{tare,gb_planner,bircher2016receding} and a line of learned policies that reason over the graph with Graph Neural Networks \citep{cao2023ariadnereinforcementlearningapproach,cao2024deep,zhan2025grate}. Despite this, no shared tool builds and maintains these graphs. Deployment planners construct them inside planner-specific, single-robot code. Learning pipelines rebuild them for every environment on the CPU, so a 16-worker training run spends much of its wall-clock time maintaining graphs rather than learning. GPU graph libraries \citep{fey2019pyg,cugraph} target fixed graphs consumed by a model, not graphs that grow and are merged and re-scored at every step of a rollout.

\begin{figure}[t]
    \centering
    \includegraphics[width=\columnwidth]{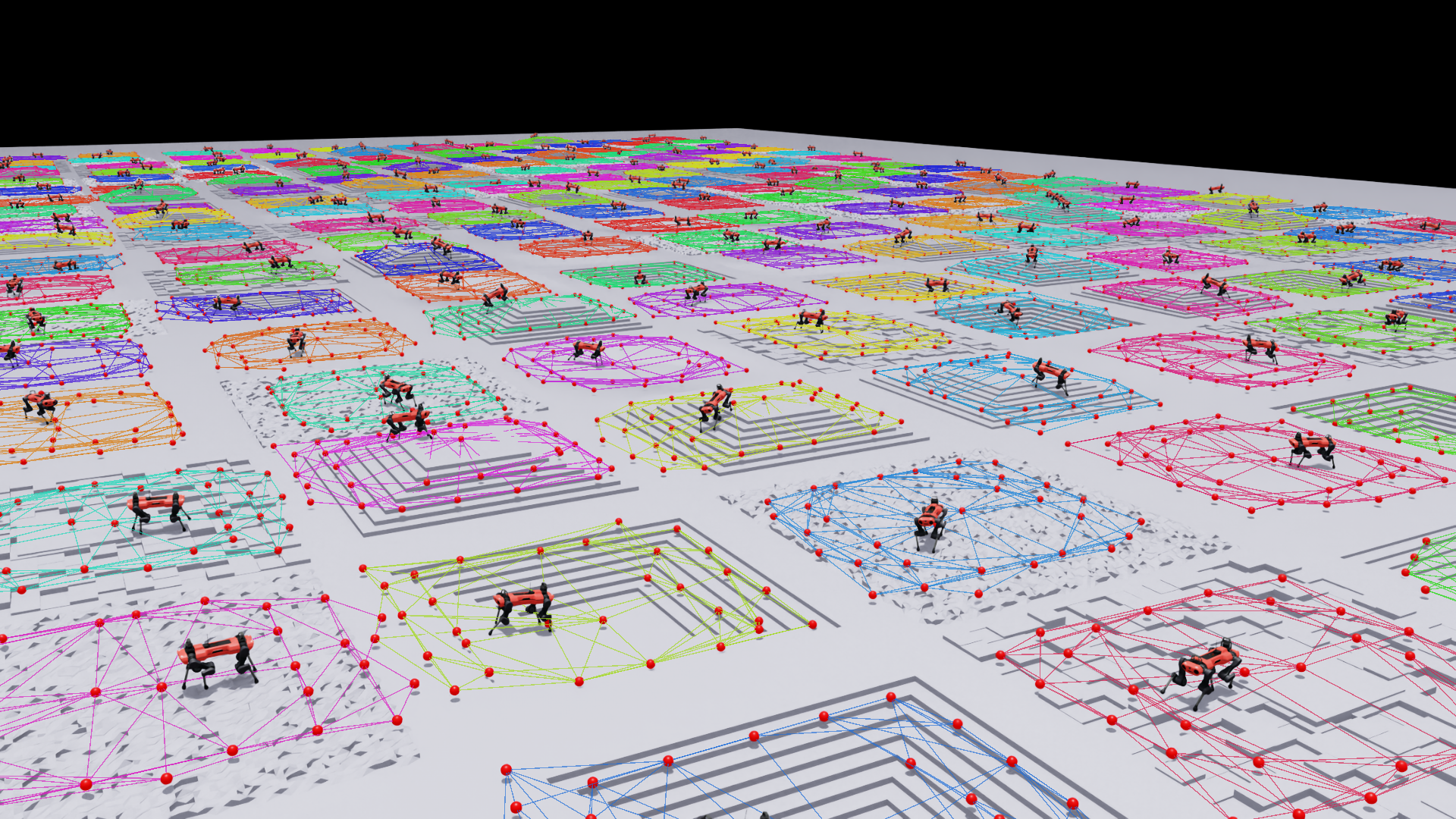}
    \caption{\small\textbf{\sysname is a GPU-accelerated graph library for robot deployment and learning.} The image shows 200 robots training graph-based exploration policies in parallel in Isaac Sim. Nodes (red) are sampled in 3-D free space and edges (colored per robot) show collision-free connectivity.}
    \label{fig:teaser}
\end{figure}

To address this gap, we present \sysname, a GPU-resident navigation-graph library that serves as one design tool for both deployment and learning (Fig.~\ref{fig:teaser}). \sysname takes occupancy grids or elevation maps as input. It samples collision-free waypoints, merges them incrementally into a persistent world-frame graph with stable node identities, and attaches per-node attributes through a layer system. Shortest paths, connectivity and frontiers are all available as CUDA tensors. The same implementation serves single-robot deployment through ROS~2 and policy learning across many parallel environments.

Building such a library raises some technical challenges. First, the graph changes at every step, as each observation adds a varying number of nodes and edges that must be merged into a graph already covering the explored world. Second, existing graph builders rely on sequential data structures, such as priority queues and per-node objects, that cannot run in parallel on a GPU. Third, learning requires hundreds of such graphs, each growing at a different rate, to be updated together. Fourth, on a real robot the graph must survive loop closure, which moves every node placed since the drift began. \sysname stores each graph as parallel tensors with stable node identities, so adding, merging, pruning and correcting nodes are single vectorized operations. It merges each observation with one batched nearest-neighbor assignment and one line-of-sight kernel over all candidate edges. It also reformulates graph search as edge-parallel Bellman--Ford and pointer-jumping connected components, running one robot or $N$ environments through the same kernels. The novel contribution of \sysname is to make the navigation graph itself a batched, GPU-resident, loop-closure-aware data structure shared by deployment and learning. To our knowledge, no other library provides this capability. Our contributions are:

\begin{enumerate}
\item \textbf{An open-source GPU navigation-graph library.} An
extensible layer system stores geometric, semantic and learned
attributes per node, and ROS~2 and batched interfaces let the same code
serve deployment and learning (Sec.~\ref{sec:design}).
\item \textbf{Validation in robot deployment and policy learning.} We use \sysname to build 3-D graphs from real-world sensor data, attach frontier and semantic layers, and correct the whole graph after loop closure in one vectorized pass. Used as a batched RL environment, \sysname trains graph-based exploration policies that reach the quality of prior learned policies in a fraction of the training time and outperform baselines on held-out maps, on a ground vehicle in a high-fidelity simulator, and on a real robot (Secs.~\ref{sec:deployment} and~\ref{sec:learning}).
\item \textbf{Scalability benchmarks.} Graph updates take just tens of milliseconds as graphs grow to a million nodes, and environment-step throughput scales with the number of parallel robots, reaching 27$\times$ that of a SoTA baseline (Secs.~\ref{sec:update_time} and~\ref{sec:env_throughput}).
\end{enumerate}

\section{Related Work}
\label{sec:related_works}
We review representations for navigation, graph software libraries, and learning-based exploration, identifying in each the gap that \sysname addresses.

\subsection{Representations for Navigation}
Early navigation in unknown environments used 2-D occupancy grids \citep{elfes1989using}, with frontier-based exploration driving the robot toward free cells that border unexplored space \citep{yamauchi1997frontier}. Volumetric maps such as OctoMap \citep{hornung2013octomap} and Voxblox \citep{Oleynikova_2017} store occupancy or distance to surfaces across a 3-D volume. Topological maps \citep{kuipers1991robot,thrun1998learning,choset2000sensor} keep only the places a robot can reach and the transitions between them, and scene graphs \citep{hughes2022hydra} organize places into a semantic hierarchy. Exploration planners combine both. GBPlanner \citep{gb_planner} plans over local and global graphs built on a volumetric map, and TARE \citep{tare} keeps a sparse roadmap along the past trajectory with dense point-cloud data around the robot. In both, the graph sits on a dense map that the planner still queries to score exploration gain. \sysname instead makes the graph the primary representation, storing frontier, semantic and learned attributes on its nodes and using dense maps only as transient inputs, so one graph serves both classical planners and learned policies.

\subsection{Graph Libraries and Ecosystems}
Navigation planners mutate their graphs as the robot explores, merging local observations into a global graph in custom C++ or Python code inside each planner \citep{tare,gb_planner,cao2023ariadnereinforcementlearningapproach}. SWAGGER \citep{swagger} builds sparse waypoint graphs from occupancy grid maps using NetworkX \citep{SciPyProceedings_11} with partial GPU support, but derives each graph from a single map rather than merging across observations. Outside navigation, GPU graph libraries serve other workloads. Graph learning frameworks such as PyTorch Geometric \citep{fey2019pyg} are fully tensorized and batch graphs by stacking adjacency matrices block-diagonally, but treat each graph as a fixed model input without spatial merging, and graph analytics libraries such as cuGraph \citep{cugraph} run GPU analytics over graphs built from edge lists. As Table~\ref{tab:method_comparison} summarizes, navigation tools are largely CPU-based and GPU libraries lack spatial merging and navigation operations. \sysname combines GPU acceleration, spatial merging, navigation and exploration operations, and batched policy learning in one library.

\subsection{Learning-Based Robot Exploration}
Learning-based exploration trains reinforcement learning policies to choose where a robot goes next, and graphs suit them because one graph is both the map the robot traverses and the input a graph neural network \citep{velickovic2018gat} reasons over. Chen et al.\ \citep{chen2020autonomousexplorationuncertaintydeep} trained graph neural network policies with DQN and A2C over a graph of past poses, landmarks and candidate frontiers for exploration under localization uncertainty. ARiADNE \citep{cao2023ariadnereinforcementlearningapproach} connects uniformly distributed viewpoints into a collision-free graph and trains an attention network with soft actor-critic to pick the next neighboring viewpoint, using each node's count of observable frontiers as a feature. Its large-scale successor \citep{cao2024deep} adds graph rarefaction and a privileged critic, and GRATE \citep{zhan2025grate} uses a graph transformer encoder and Kalman-filtered waypoints. These three train in 2-D grid worlds with 16 to 32 Ray workers for one to three days, and in ARiADNE's released code each worker builds its graph in Python on the CPU. \sysname builds and merges every environment's graph on the GPU, so hundreds of environments step as one tensor operation, which makes on-policy methods such as PPO \citep{schulman2017ppo} practical, as in massively parallel locomotion learning \citep{rudin2022walk}.

\begin{table}[t]
    \centering
    \begin{threeparttable}
        \caption{Comparison of graph libraries and planners across the capabilities a navigation graph needs for deployment and learning.}
        \label{tab:method_comparison}
        \setlength{\tabcolsep}{3.5pt}
        \scriptsize
        \begin{tabular}{lcccc}
            \toprule
            \textbf{Library} &
            \makecell{\textbf{GPU} \\ \textbf{Accel.}} &
            \makecell{\textbf{Spatial} \\ \textbf{Graph} \\ \textbf{Merging}} &
            \makecell{\textbf{Robotics} \\ \textbf{Navigation \&} \\ \textbf{Exploration}} &
            \makecell{\textbf{Policy} \\ \textbf{Learning} \\ \textbf{Support}} \\
            \midrule
            NetworkX           & \xmark & \xmark & \xmark & \xmark \\
            cuGraph \citep{cugraph}           & \cmark & \xmark & \xmark & \xmark \\
            PyTorch Geometric \citep{fey2019pyg} & \cmark & \xmark & \xmark & \cmark \\
            GBPlanner \citep{gb_planner}         & \xmark & \cmark & \cmark & \xmark \\
            TARE Planner \citep{tare}      & \xmark & \cmark & \cmark & \xmark \\
            ARiADNE node manager \citep{cao2023ariadnereinforcementlearningapproach} & \xmark & \cmark & \cmark & \otick \\
            SWAGGER \citep{swagger}           & \otick & \xmark & \cmark & \xmark \\
            \midrule
            \sysname (ours)         & \cmark & \cmark\tnote{*} & \cmark & \cmark \\
            \bottomrule
        \end{tabular}
        \begin{tablenotes}
            \tiny\raggedright
            \item \otick~partial support (mainly CPU based); *~batched across parallel environments.
        \end{tablenotes}
    \end{threeparttable}
\end{table}

\section{Problem Formulation}
\label{sec:problem}
To showcase accelerated training for robot exploration policies that take graphs as inputs, we consider the problem of a ground robot in an a priori unknown environment whose task is to explore and observe all reachable free space while
traveling the shortest possible path. At each timestep $t$ the robot
has pose $\mathbf{x}_t \in \mathrm{SE}(3)$ and receives a local
observation $\mathcal{O}_t = \{\mathbf{M}_t, \mathbf{x}_t\}$, where
$\mathbf{M}_t$ is a dense representation of the local environment centered on
$\mathbf{x}_t$.
Our choice of representation for the entire environment is a graph that both classical planners 
and learned policies act on. \sysname maintains this graph incrementally as a stitched,
world-frame graph
$\mathcal{G}_t = (\mathcal{V}_t, \mathcal{E}_t, \mathcal{S}_t)
= f(\mathcal{G}_{t-1}, \mathcal{O}_t)$, where $\mathcal{V}_t$ are
collision-free waypoints with positions
$\mathbf{p}_i \in \mathbb{R}^3$, $\mathcal{E}_t$ are collision-free
edges consistent with observations
$\{\mathcal{O}_\tau\}_{t_{0} \le \tau \le t}$, and
$\mathcal{S}_t : \mathcal{V}_t \to \mathbb{R}^L$ is an $L$-channel score
map that attaches geometric, semantic, or learned attributes to each
node. At each step the robot sits at a node $u_t \in \mathcal{V}_t$ and must
decide which adjacent node to travel to next. We formulate this
decision as a Markov decision process (MDP) over $\mathcal{G}_t$. The action is a node
$a_t = v \in \mathcal{V}_t$ with $(u_t, v) \in \mathcal{E}_t$, which the
robot drives to. The transition is
$s_{t+1} = [\,f(\mathcal{G}_t, \mathcal{O}_{t+1}),\, \mathbf{x}_{t+1}\,]$.
The reward $r_t$ credits newly observed free space and penalizes
distance traveled. We seek a policy $\pi$ from
which the action is sampled, $a_t \sim \pi(\cdot \mid s_t)$, and the
objective is
$\pi^* = \arg\max_{\pi} \mathbb{E}_{\pi}\!\left[\sum_{t=1}^{T} r_t\right]$.
\section{Methodology}
Our approach has two components. A \emph{batched graph builder} (Sec.~\ref{sec:design}) maintains the
navigation graph. \sysname implements the update $f$ of
Sec.~\ref{sec:problem} as a GPU pipeline that samples a local graph from each
observation $\mathcal{O}_t$, merges it into the persistent world-frame graph
$\mathcal{G}_t$, and writes the per-node scores $\mathcal{S}_t$, running one
robot or $N$ environments through the same kernels. A \emph{parallel
reinforcement-learning pipeline} (Sec.~\ref{sec:meth_exploration}) then solves the MDP of
Sec.~\ref{sec:problem}. We parameterize the policy $\pi$ as a Graph Neural
Network (GNN) that acts directly on $\mathcal{G}_t$ and train it across
hundreds of \sysname environments stepped together on the GPU.
\subsection{Batched Graph Builder}
\label{sec:design}

\textbf{Local graph construction.} A frontier kernel scans the input map once, emitting the free-space mask and the frontier set in a single pass. The waypoint sampler then populates the mask with well-spaced nodes through three complementary passes. A \emph{skeleton} pass places nodes along the medial axis of free space, a \emph{boundary} pass places nodes along inflated obstacle contours to cover corridors and walls, and a \emph{free-space} pass uses a GPU distance transform to fill open regions until every free cell lies within a fixed radius of some node. This is a tensorized implementation of the graph builder logic in NVIDIA's SWAGGER library \citep{swagger}. For learning experiments that require the uniform lattice of prior work \citep{cao2023ariadnereinforcementlearningapproach}, a world-anchored lattice can replace the three passes without changing anything downstream. For 2.5-D elevation inputs, a slope and step-height stage lifts the map into a 3-D traversability volume before sampling, which is what produces the graphs over ramps and stairs.

\begin{figure}[t]
    \centering
    \includegraphics[page=1, width=\columnwidth, trim=1pt 0pt 1pt 1pt, clip]{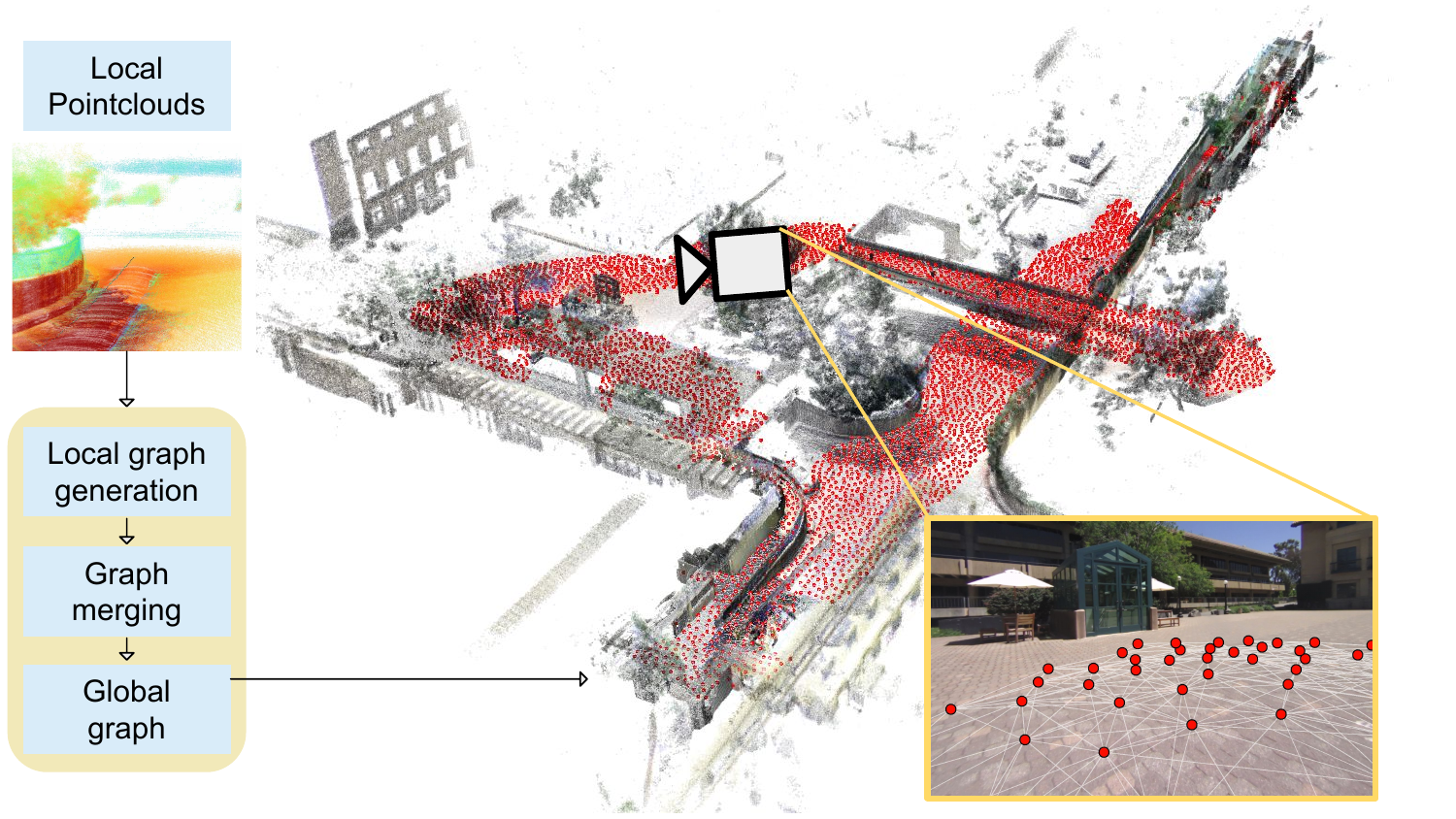}
    \caption{\small\textbf{\sysname constructs 3-D navigation graphs from real-world data captured with an Odin camera.} Local point clouds are converted into occupancy grids, from which the library samples nodes (red). The inset shows the graph projected into the camera view. The traversable structure of a multi-level outdoor scene is stored in a few thousand nodes, whereas a volumetric map of the same scene would require orders of magnitude more memory.}
    \label{fig:odin}
\end{figure}

\textbf{Global graph assembly.} Each frame's local graph is merged into the global graph by a single batched nearest-neighbor assignment, in which every local node is either snapped to an existing global node or appended with a fresh, unique identifier. Candidate edges are proposed by a radius query and validated with GPU-parallel line-of-sight checks against the obstacle memory, with surviving edges added subject to a per-node degree cap and a periodic pruning of redundant edges. Node and edge arrays are parallel tensors, so every mutation is a vectorized gather or scatter, and a retention rule prunes long-unobserved nodes to bound graph size. Fig.~\ref{fig:odin} showcases a real-world implementation of \sysname building a 3-D graph in a campus setting with bridges, slopes, ramps and underpasses. 

\textbf{Multi-layer attribute system.} Per-node attributes are exposed through registered \emph{graph layers}. A \emph{compute layer} executes a user kernel against the current graph each frame (the built-in frontier layer of Fig.~\ref{fig:layers}A is one), while an \emph{external layer} ingests asynchronous scores from outside the library and persists them across frames (the terrain-semantics layer of Fig.~\ref{fig:layers}B). Merge policies (overwrite, blend, moving average) govern how repeated observations combine. Every layer is a GPU tensor aligned with the node array, so its output is consumed without copies, as a feature matrix for a policy or a scalar score for a planner.

\textbf{GPU graph search.} For planning, the global graph is materialized in compressed-sparse-row form directly from the edge tensors. \sysname provides parallel breadth-first search, a Bellman--Ford kernel for single- and multi-source shortest paths, and pointer-jumping connected components. Bellman--Ford is preferred over Dijkstra because it relaxes all edges in one kernel launch, whereas a priority queue serializes.

\textbf{Loop closure.} Each node stores the index of the odometry frame at which it was created. When an external SLAM module supplies a corrected trajectory, graph correction reduces to a single $O(N)$ vectorised gather that translates each node by the displacement of its parent frame, with no graph traversal, no sensor re-integration, and no occupancy reprojection. A subsequent GPU node remerge collapses spatially coincident nodes from overlapping traversals, and a frontier validity recheck demotes boundary nodes that the correction has rendered interior to the graph's free nodes.

\textbf{Batched execution.} For policy training, $N$ independent builders share the same kernels and are stepped together each simulator tick, producing padded position, adjacency, frontier, and attribute tensors that a graph-conditioned policy consumes directly. Because per-environment graph state is itself GPU-resident, only kernel saturation, not Python or PCIe overhead, bounds the batch size (Sec.~\ref{sec:env_throughput}).

\textbf{ROS~2 deployment.} A ROS~2 wrapper accumulates a sensor stream and odometry into a rolling map, feeds it to the incremental builder, and exposes the layer system so external modules publish per-position scores. The exploration node of Sec.~\ref{sec:sim_deploy} is one such wrapper.

\subsection{Parallel Policy Learning For Exploration}
\label{sec:meth_exploration}
In order to utilize \sysname's batched graph builder, we train
  our GNN exploration policy using SAC or PPO~\citep{schulman2017ppo},
  following the massively parallel training setup
  of~\citep{rudin2022walk}. Unlike previous work where SAC was used \citep{cao2023ariadnereinforcementlearningapproach, cao2024deep, zhan2025grate}, \sysname unlocks the use of on-policy PPO and large-scale parallelization by providing high-throughput environment data. \sysname maintains a navigation graph
  for each of $N$ environments. We use separate actor
  and critic networks, each shared across all environments. The entire training loop runs on the GPU, including graph
  maintenance, environment updates, observation and reward computation. Keeping the environment and learner on the GPU avoids CPU--GPU
  transfers.

  \textbf{Exploration environment:}
  \label{sec:exploration_env}
  We train exploration policies in 2-D grid worlds represented as
  occupancy maps. Each map specifies free and occupied cells, while
  the robot maintains a belief map that additionally marks unobserved
  cells as unknown. Simulated sensor observations update this belief
  as the robot moves, and \sysname maintains a navigation graph
  from the observed free space. We use the same training maps as ARiADNE~\citep{cao2023ariadnereinforcementlearningapproach},
  obtained from its public repository. Since ARiADNE does not
  release its validation split, we generate 150 new maps using
  the dungeon generator of~\citep{chen2019self}. These new validation maps are
  excluded from training and grouped into easy, medium, and complex
  tiers of 50 maps each, with difficulty increasing through the
  number of dead-end corridors and the free-space
  boundary-to-area ratio.

  \textbf{Reward:}
  We use the exploration reward of prior
  work~\citep{cao2023ariadnereinforcementlearningapproach,cao2024deep,zhan2025grate},
  which rewards exploration progress and penalizes travel distance.
  A frontier cell is an unknown cell adjacent to known free space.
  At each step, the robot receives reward $r_t = a \cdot r_f - b \cdot r_d + r_c,$ where $r_f = F(\mathcal{G}_{t-1}, \mathcal{G}_t)$ counts the frontier
  cells removed during the step and
  $r_d = \|\mathbf{x}_t - \mathbf{x}_{t-1}\|_2$ is the distance
  traveled. We use the same coefficients as \cite{cao2023ariadnereinforcementlearningapproach}, $a = 1/64$ and $b = 1/50$,
  with a completion bonus of $r_c = 20$ when exploration is
  completed and $r_c = 0$ otherwise.

  \textbf{Training details:}
  \label{sec:training_details}
  Each environment maintains an independent exploration episode. At each PPO iteration, we hold the policy
  parameters fixed and collect $T$ steps from each of $N$ environments,
  producing a rollout batch of
      $B = NT$
  transitions. We then update the actor and critic using mini-batches
  drawn from this rollout. After optimization, collection resumes from the current environment
  states using the updated policy. Environments reset independently
  upon episode termination or reaching the episode-length limit.
 We select the rollout batch size $B$ and its allocation between
  $N$ parallel environments and $T$ steps per environment through
  the wall-clock training experiments in Sec.~\ref{sec:training_time}.
  Actor and critic each use ARiADNE's attention-based encoder--decoder~\citep{cao2023ariadnereinforcementlearningapproach} (6 layers, 8 heads, 128-d embeddings). PPO runs 4 epochs of 4 mini-batches per iteration with clip 0.2, $\gamma = 1$, $\lambda = 0.95$, entropy coefficient 0.01, and an adaptive learning rate (initially $10^{-4}$, KL target 0.01); SAC keeps ARiADNE's hyperparameters.

\section{Experiments}
We demonstrate the capabilities of \sysname by answering the following questions:  (Q1) How does \sysname compare to existing open-source libraries for single-robot deployment? (Q2) Can \sysname accelerate robot policy learning by maintaining a separate graph for each robot across thousands of parallel environments? (Q3) Can policies trained using \sysname control robots to explore unknown environments in high-fidelity simulators and the real world? 
\subsection{Robot Deployment}
\label{sec:deployment}
\subsubsection{Real-World 3-D Graphs}
We first validate \sysname on real sensor data. Fig.~\ref{fig:odin} shows the library running on a sequence captured by a Boston Dynamics Spot robot with a Manifold Odin1 LiDAR-camera mounted on it, traversing a university campus. Local point clouds are processed by EM-CuPy~\cite{miki2022elevation} into elevation maps $\mathbf{M}_t$, from which \sysname samples a local graph. Applying $f$ at each step merges these into a single world-frame graph $\mathcal{G}_t$ spanning several elevation levels. The resulting representation is directly usable by a planner, since nodes are collision-free places, edges are traversable
transitions, and shortest paths and frontier queries execute on the GPU. 
\begin{figure}[t]
    \centering
    \includegraphics[page=1, width=\columnwidth, trim=2pt 120pt 2pt 1pt, clip]{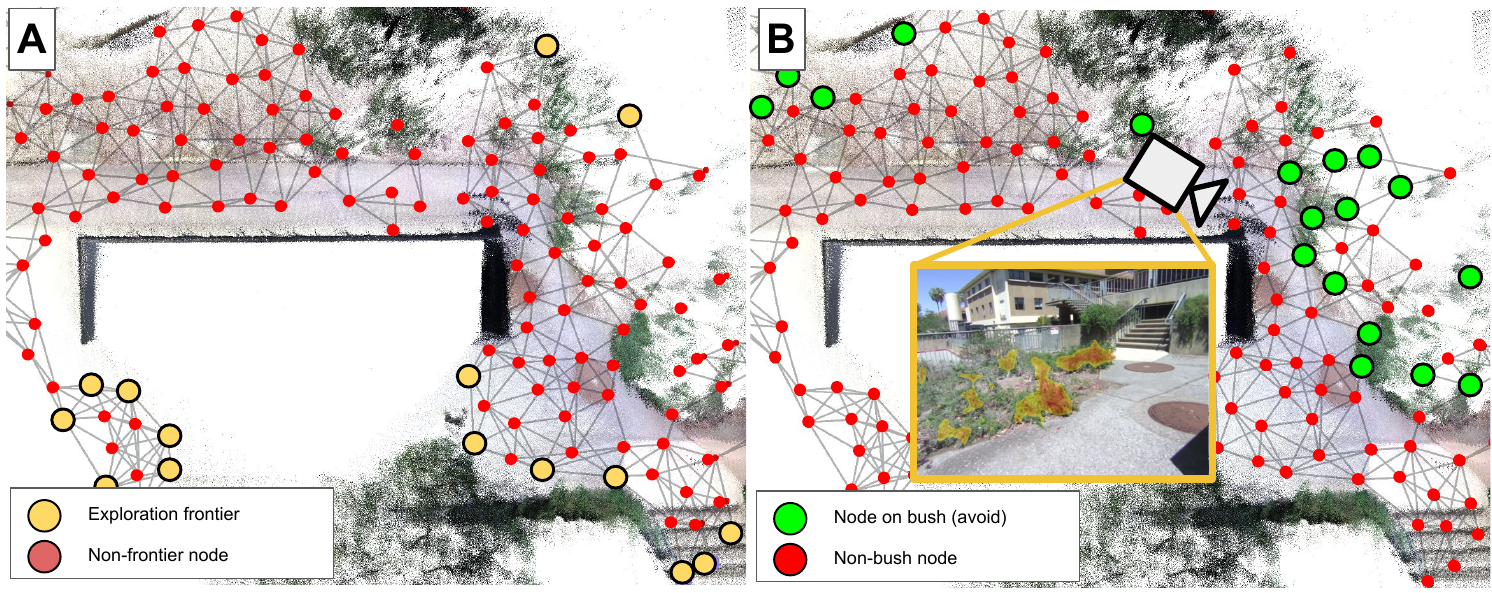}
    \caption{\small\textbf{Node attributes for exploration and navigation.} Bird's-eye view of one region of the campus graph of Fig.~\ref{fig:odin}. \textbf{(A)} The frontier layer marks nodes on the boundary between explored and unexplored space (yellow) against free-space nodes (red). \textbf{(B)} A semantic layer marks nodes on grass or bushes (green) against paved ground (red), using RGB input.}
    \label{fig:layers}
\end{figure}
Fig.~\ref{fig:layers} shows channels of the score map $\mathcal{S}_t$. The
frontier channel (Fig.~\ref{fig:layers}A) is computed internally. At
each step, nodes adjacent to the known-free/unknown boundary are
flagged, and this channel is what the policies of Sec.~\ref{sec:learning} consume. The semantic
channel (Fig.~\ref{fig:layers}B) is supplied externally. Nodes are
scored by ExploRFM~\cite{shah2026wildos}, a foundation-model-based
vision module, which can label them as grass or pavement from the camera
view, and $f$ accumulates repeated observations so the label stabilizes
as the robot moves. 
\subsubsection{Loop Closure}
\label{sec:loop}
\begin{figure}[t]
    \centering
    \renewcommand{\thesubfigure}{\Alph{subfigure}}
    \begin{subfigure}[b]{0.49\columnwidth}
        \centering
        \includegraphics[width=\textwidth, trim=10pt 0pt 637pt 0pt, clip]{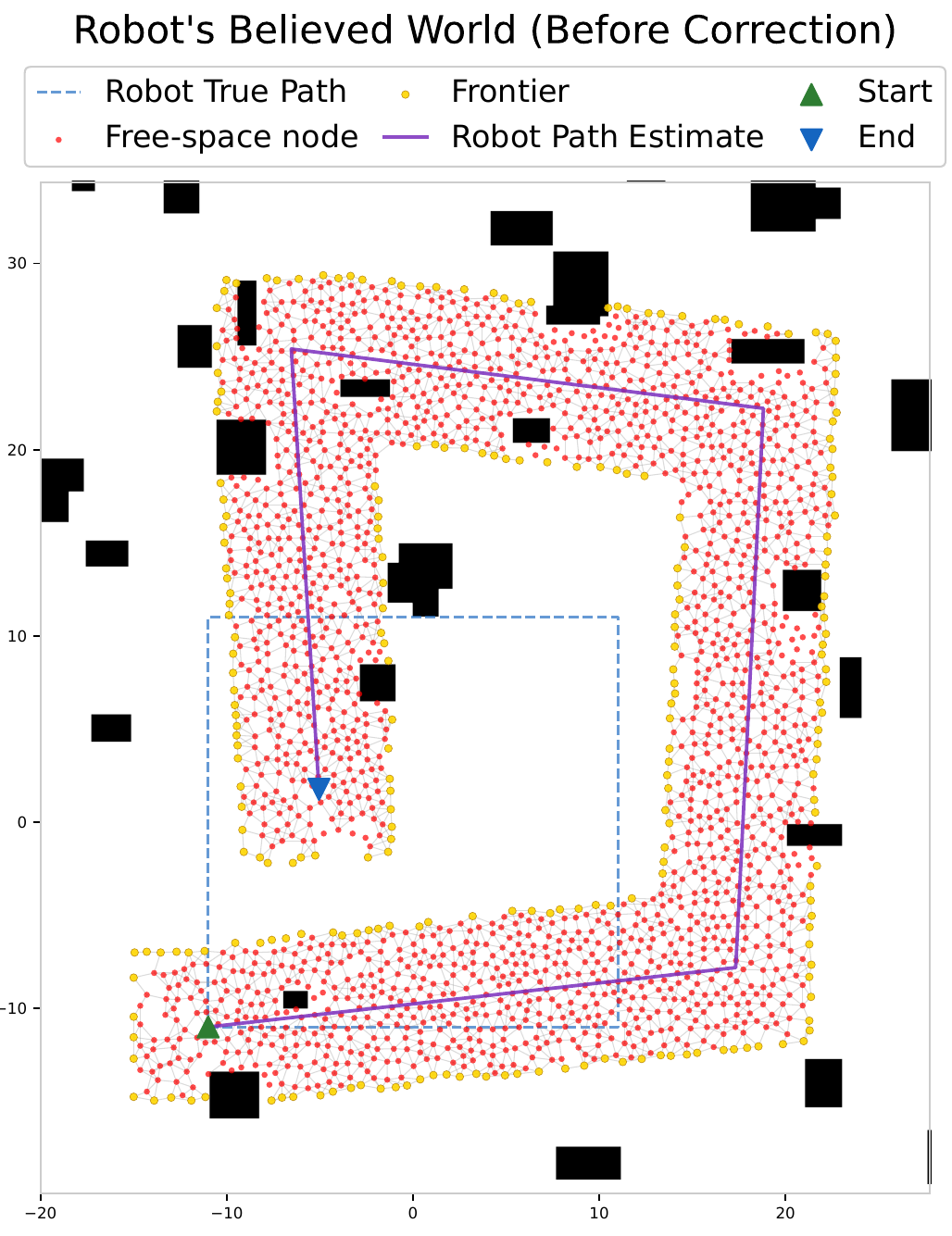}
        \caption{}
        \label{fig:loop_before}
    \end{subfigure}
    \hfill
    \begin{subfigure}[b]{0.49\columnwidth}
        \centering
        \includegraphics[width=\textwidth, trim=647pt 0pt 0pt 0pt, clip]{images/odom_correction_figs_new_8_rect_scale22pct_add160mmpm_believed_vs_corrected_v2.pdf}
        \caption{}
        \label{fig:loop_after}
    \end{subfigure}
    \caption{\small\textbf{Loop closure on a \sysname graph.} The graph of a drifted square trajectory \textbf{(A)} before and \textbf{(B)} after the correction. Dashed: true path; solid: estimated path.}
    \label{fig:loop}
\end{figure}
We evaluate on a closed square trajectory under a combined drift of 22\,\% scale error and 160\,mm/m
 additive world-frame bias, producing a pre-correction loop-closure gap of 14.08\,m (Fig.~\ref{fig:loop_before}). The full correction pipeline restores the closed loop (Fig.~\ref{fig:loop_after}) in 2.26\,ms. Such efficiency is not feasible for volumetric representations. An occupancy grid or TSDF over the same environment requires an $O(W{\times}H)$ pixel-level warp followed by re-ray-casting or signed-distance fusion. In \sysname, each node carries the index of the odometry step that placed it, so loop closure reduces to a single vectorised pass.

\subsubsection{Graph Update Time}
\label{sec:update_time}
\begin{figure}[t]
    \centering
    \includegraphics[width=0.9\columnwidth, trim=7pt 0pt 0pt 0pt, clip]{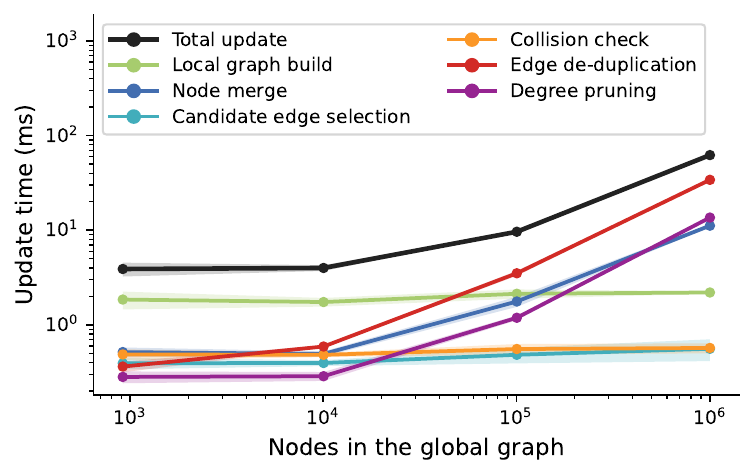}
    \caption{\small\textbf{Graph update time vs number of nodes.} Each point is one \sysname update. A local graph built from an 80$\times$80\,m observed window is merged into a global graph of $N$ nodes, with collision checks on the new edges. Lines show the total and per-stage update time, averaged over three obstacle layouts.}
    \label{fig:update_scaling}
\end{figure}

A deployed planner updates its graph every cycle, so the cost of that update bounds how large a world a single robot can map in real time. Fig.~\ref{fig:update_scaling} measures \sysname update time as the global graph grows. A full update takes about 7\,ms at $100{,}000$ nodes. Compared to a SoTA baseline implementation using SWAGGER and NetworkX, this is about 120$\times$ faster. The update time is just 62\,ms at a million nodes. At a 0.5\,m node sampling distance, a million nodes can cover a 500$\times$500\,m area. Building the local graph, selecting candidate edges and checking collisions take about 3\,ms at every size, and the growth comes from merging nodes, de-duplicating edges and pruning node degrees.

\begin{table*}[t]
    \centering
    \caption{Performance of the learned exploration policy on 150 validation maps (50 per difficulty tier).
    Values are mean path length in grid cells (px). Success is the fraction of maps in which complete coverage was achieved.}
    \label{tab:gridworld}
    \setlength{\tabcolsep}{5pt}
    \footnotesize
    \begin{tabular}{llccccc}
        \toprule
        \textbf{Planner} & \textbf{Graph builder} & \textbf{Easy} & \textbf{Medium} & \textbf{Complex} & \textbf{All} & \textbf{Success rate} \\
        \midrule
        Nearest frontier (greedy) & ARiADNE & 1404 & 1896 & 2388 & 1896 & 0.44 \\
        ARiADNE SAC-16 (6\,h) & ARiADNE & 909 & 1304 & 2305 & 1506 & 0.86 \\
        ARiADNE released policy (24\,h, 32 robots) & ARiADNE & 861 & \textbf{1090} & 2050 & 1333 & 0.91 \\
        SAC-512 (6\,h) & \sysname & \textbf{736} & 1205 & 2165 & 1369 & \textbf{0.98} \\
        PPO ($512 \times 32$, 6\,h) & \sysname & 785 & 1156 & \textbf{1825} & \textbf{1255} & 0.97 \\
        \bottomrule
    \end{tabular}
\end{table*}

\subsection{Robot Learning}
\label{sec:learning}
Autonomous exploration underpins search and rescue and inspection, and
graph-based planners have been fielded in settings like the DARPA
Subterranean Challenge \citep{gb_planner,tare}. Learned policies over the same
graphs now report the shortest paths on common benchmarks~\citep{cao2023ariadnereinforcementlearningapproach,cao2024deep,zhan2025grate},
but their training infrastructure has not kept pace. Massively parallel GPU
simulation cut locomotion training from days to minutes \citep{rudin2022walk},
while graph-based exploration still rebuilds its environment on the CPU.
 Thus, \sysname's robot learning experiments are aimed at showing how the library serves as a foundation for accelerating modern exploration policy training. We build on prior work on attention-based graph exploration
policies~\citep{cao2023ariadnereinforcementlearningapproach,cao2024deep,zhan2025grate} where robots explore 2-D grid worlds while maintaining graphs in Python. In these methods, experience is collected in parallel using Ray~\citep{moritz2018raydistributedframeworkemerging} at one worker per environment, capping parallelism at CPU thread count. We
replicate their experiments using \sysname to maintain the navigation
graph for each robot, then measure the resulting gains in throughput,
training time and final policy performance.
\subsubsection{Batched Environment-Step Throughput}
\label{sec:env_throughput}
We compare environment-step throughput between \sysname and ARiADNE's CPU-based graph manager as the number of parallel robots grows. A step applies the selected action, updates the graph, and returns the next observation and reward.

\textbf{Setup:} Both pipelines run ARiADNE's released
checkpoint on 200
maps drawn at random from the training set of
Sec.~\ref{sec:exploration_env}, with identical graph-sampling parameters
so each measurement spans the same distribution of graph sizes. The
measured interval begins at the end of policy inference and closes once
the next observation, reward and termination flags are available. For $N$ parallel robots we accumulate wall-clock time
$T_{\mathrm{env}}$ over $H = 128$ steps per robot and report throughput
$R_{\mathrm{env}} = NH/T_{\mathrm{env}}$ in environment steps per
second. \sysname batches updates across $N$ environments on the GPU
while ARiADNE distributes them over 16 Ray workers. We sweep $N$ from $1$
to $1024$ on an AMD Ryzen~7 8700F with an RTX 5070~Ti.

\begin{figure}[t]
    \centering
    \renewcommand{\thesubfigure}{\Alph{subfigure}}
    \begin{subfigure}[b]{0.49\columnwidth}
        \centering
        \includegraphics[width=\textwidth, trim=10pt 0pt 0pt 0pt, clip]{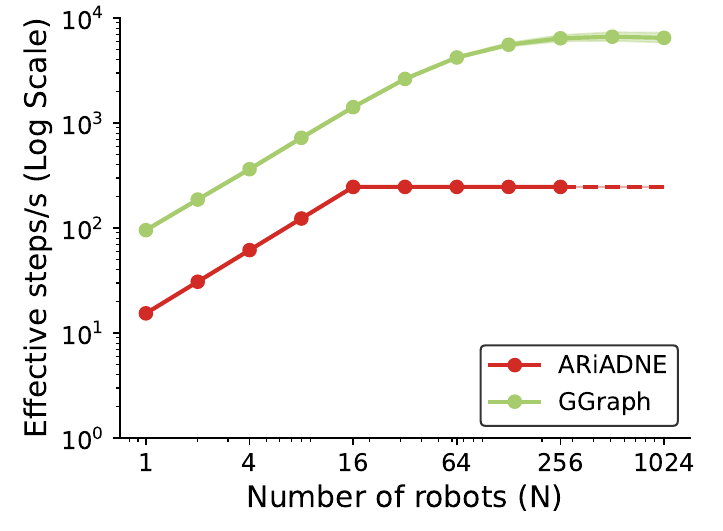}
        \caption{}
        \label{fig:per_step_env_batch}
    \end{subfigure}
    \hfill
    \begin{subfigure}[b]{0.49\columnwidth}
        \centering
        \includegraphics[width=\textwidth, trim=7pt 0pt 0pt 0pt, clip]{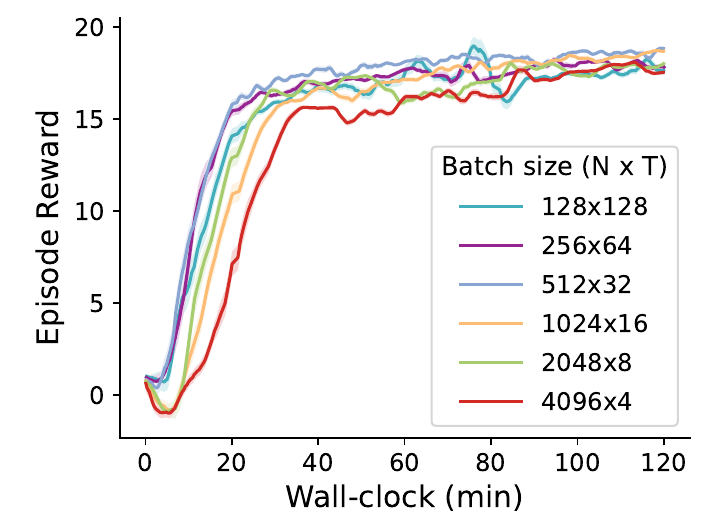}
        \caption{}
        \label{fig:sweep}
    \end{subfigure}
    \caption{\small\textbf{Throughput and rollout allocation.}
    \textbf{(A)} Complete environment-update throughput for \sysname and
    ARiADNE as the number of parallel robots increases; both axes are
    logarithmic. \textbf{(B)} PPO reward versus training wall-clock time.
    Each configuration collects $NT = 16{,}384$ transitions per
    iteration, with $N$ environments and $T$ steps per environment.}
    \label{fig:throughput_sweep}
\end{figure}

\textbf{Results:} Fig.~\ref{fig:per_step_env_batch} shows that the
throughput advantage of \sysname grows with the number of parallel
robots, from approximately 6$\times$ at $N = 16$
to 26$\times$ at $N = 256$. Its throughput then levels off, reaching
about 6{,}500 env-steps per second at $N = 512$, compared with 244 for ARiADNE, a 27$\times$ speedup. 

\subsubsection{Wall-Clock Training Efficiency and Performance on 2-D Grid Worlds}
\label{sec:training_time}
We evaluate how faster environment updates affect the wall-clock time
required to learn an exploration policy. We first examine how the PPO
rollout budget should be split across parallel environments, then
compare policy quality at a matched training budget. 

\textbf{Setup:}  We follow the training setup and use the training and
validation sets of Sec.~\ref{sec:training_details}.  Environment-step throughput plateaus
around $N = 256$ (Fig.~\ref{fig:per_step_env_batch}), so we first ask
how the rollout budget should be split between the number of
environments $N$ and the rollout length $T$ per environment. Holding the
batch fixed at $B = NT = 16{,}384$ transitions, we sweep $N$ from $128$
to $4096$, so $T$ falls correspondingly from $128$ to $4$ steps. We then
compare ARiADNE SAC with 16 Ray workers against \sysname-based SAC and
PPO with $512$ environments, all on an RTX~4090. Evaluation uses the 150
independently generated validation maps, 50 in each of the easy, medium
and complex tiers. Policies act greedily and an episode succeeds when no
node has positive utility within a 256-step limit.

\begin{figure}[t]
    \centering
    \includegraphics[width=0.8\columnwidth]{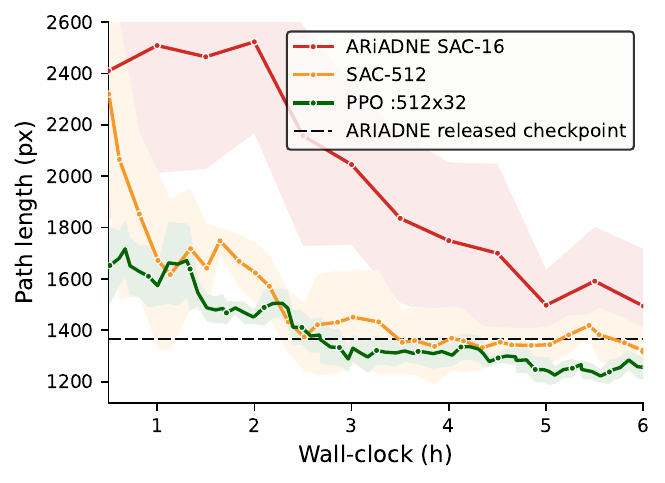}
    \caption{\small\textbf{Training curves showing exploration performance, measured by the path length required to complete exploration, versus wall-clock training time.} Lower path length is better. The dashed line marks the performance of ARiADNE’s released ckpt.
}
    \label{fig:training_time}
\end{figure}

\textbf{Results:} Fig.~\ref{fig:sweep} shows that $512 \times 32$
improves fastest early in training and holds among the highest rewards
across the two-hour run. Beyond this point the rollouts become too
short: early learning slows at $N = 2048$ and $4096$, where $T$ drops to 8 and
4 steps. Each iteration still collects the same number of transitions,
but rollouts this short truncate the return estimates the policy learns
from. Based on these results, we adopt $512 \times 32$
throughout our experiments. 

\begin{figure*}[t]
    \centering
    \includegraphics[width=0.95\textwidth]{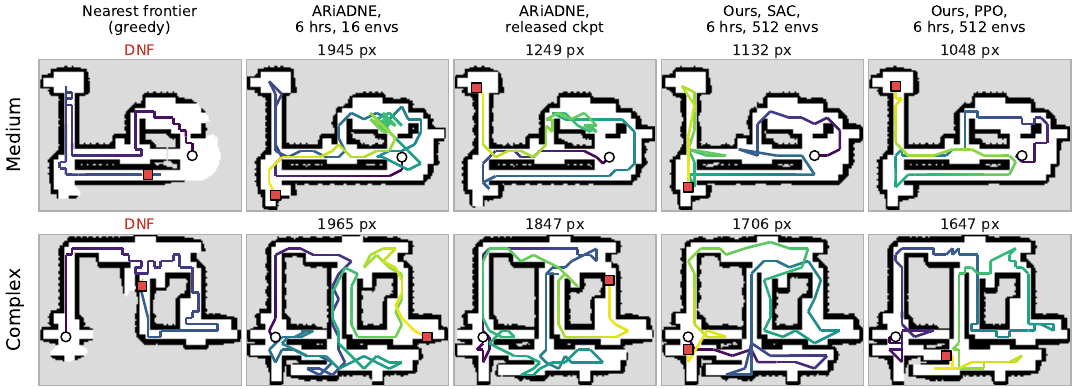}
    \caption{\small\textbf{Exploration trajectories on medium (top) and
    complex (bottom) validation maps.} PPO takes the shortest path among
    the learned policies in both examples, while the nearest-frontier
    baseline fails to complete either map. White circles mark starts, red
    squares mark final poses, and gray regions remain unknown. Labels
    report path length; DNF denotes failure to complete exploration.}
    \label{fig:gridworld_paths}
\end{figure*}

Under this configuration both \sysname pipelines learn
shorter exploration paths than ARiADNE SAC-16 within the same budget
(Fig.~\ref{fig:training_time}), with PPO falling below the path length of the released
checkpoint, trained for 24\,h, after roughly 2\,h. At the $6$\,h budget
(Table~\ref{tab:gridworld}), PPO achieves the shortest overall mean path
of $1255$\,px and completes $97$\,\% of maps, against $1333$\,px and
$91$\,\% for the released policy. The gain is largest on complex maps,
where PPO reduces mean path length from $2050$ to $1825$\,px. SAC
reaches the highest completion rate at $98$\,\% and the shortest paths
on easy maps, while the released policy remains best on medium maps.
Fig.~\ref{fig:gridworld_paths} shows the resulting performance on medium and complex maps.

\begin{figure*}[t]
    \centering
    \includegraphics[width=0.95\textwidth, trim=0pt 0pt 0pt 7pt, clip]{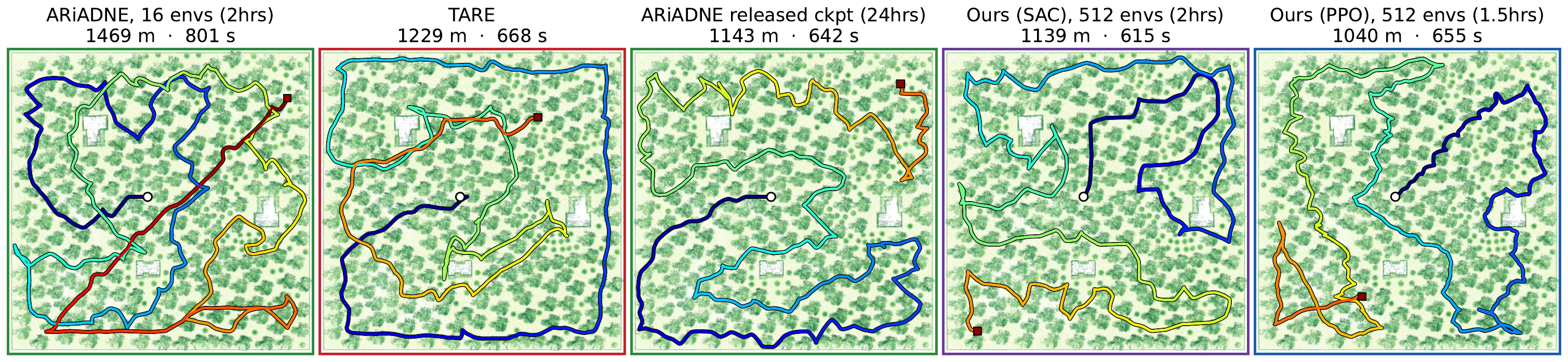}
    \caption{\small\textbf{Exploration of the \texttt{forest} world (150$\times$150\,m) by a ground vehicle in the CMU autonomous exploration development environment} \citep{cao2022autonomous}. Left to right: ARiADNE's recipe trained for 2\,h with 16 Ray workers; TARE \citep{tare}; ARiADNE's released checkpoint trained for 24\,h; our SAC policy trained for 2\,h with 512 robots on \sysname; and our PPO policy trained for 1.5\,h with 512 robots on \sysname. Each panel is a run from the same start pose, and the numbers are the distance driven and the time taken for completion. White circle: start; red square: end. As another baseline, GRATE \cite{zhan2025grate} reports a best path length of 1026\,m under its own completion rule, using a policy trained for about 36\,h. Our PPO approach on \sysname achieves similar performance in just 1.5\,h of training.}
    \label{fig:sim_deploy}
\end{figure*}

\begin{figure}[t]
    \centering
    \includegraphics[width=0.70\columnwidth, trim=1pt 0pt 1pt 40pt, clip]{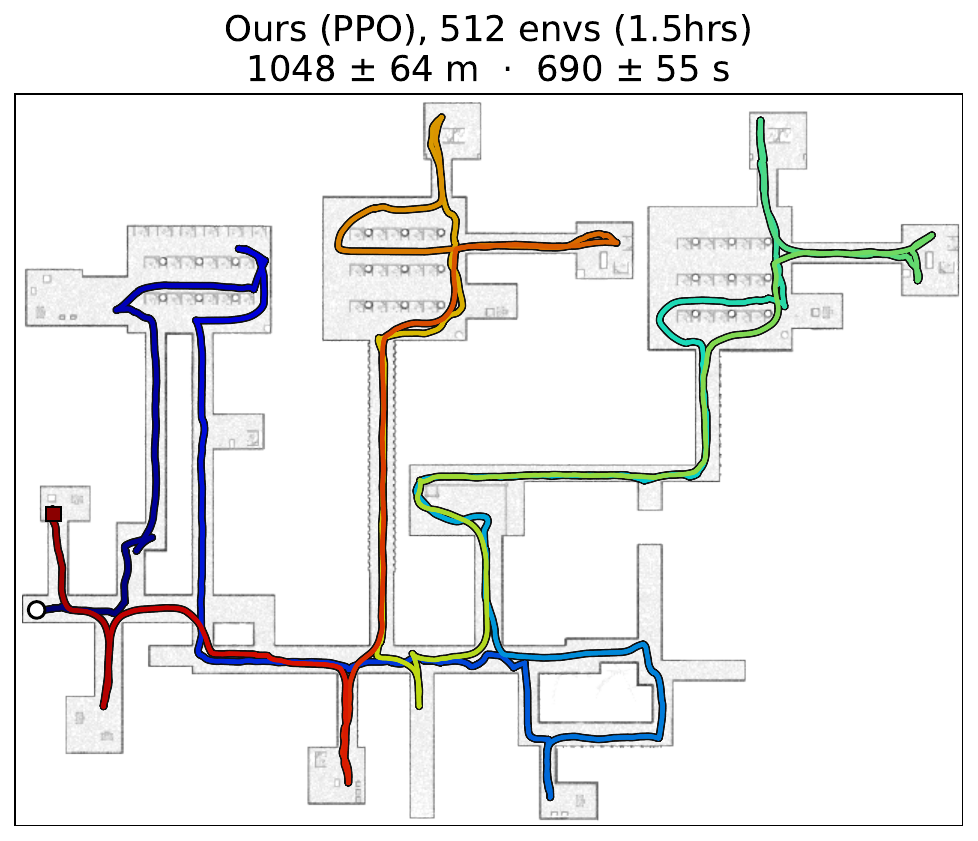}
    \caption{\small\textbf{Exploration of the large-scale indoor environment} \citep{cao2022autonomous} by our PPO-based policy trained using \sysname for 1.5\,h. It achieves a path length of $1048 \pm 64$\,m across 10 seeds. The reported baseline for exploration in the same map by ARiADNE is 1042\,m using a policy trained with SAC for 24\,h.
}
    \label{fig:indoor}
\end{figure}

\subsubsection{Deployment of the Learned Policy in a High-Fidelity Simulator and the Real World}
\label{sec:sim_deploy}

We deploy the learned policies on a simulated ground vehicle in the CMU autonomous exploration development environment \citep{cao2022autonomous}. It provides a Gazebo model of a skid-steer vehicle with a Velodyne VLP-16 LiDAR, a state estimator, terrain analysis, and a local planner that tracks a waypoint at up to 2\,m/s. An OctoMap belief built from the vehicle's registered scans is projected to a 2-D occupancy grid, a navigation graph is maintained on it, and at each step the policy selects the next goal node, which becomes the vehicle's waypoint. Each policy acts on the graph geometry it was trained on (4\,m node spacing, 16\,m sensor range, 12.8\,m utility range); ours uses the graph built by \sysname, and the ARiADNE policies use their own builder. We compare paths at equal coverage. The 150$\times$150\,m world is divided into 1\,m cells, and a cell counts as covered once the vehicle has passed within 20\,m of it. Coverage is the fraction of cells covered, and runs are cut at the moment their coverage first reaches 99.8\,\%.

Fig.~\ref{fig:sim_deploy} shows one run of each planner in a large-scale outdoor map. ARiADNE's recipe trained with its own code for 2\,h takes the longest path to explore, 1469\,m in 801\,s; much of that distance comes at the end of the run, when the vehicle is forced to cross its previous path again to reach the last unexplored regions. TARE follows at 1229\,m in 668\,s. ARiADNE's released checkpoint, reported to need about 24\,h of training, and our SAC policy, trained for 2\,h with 512 robots on \sysname, yield comparable performance. They drive 1143\,m in 642\,s and 1139\,m in 615\,s, respectively. Our PPO policy, trained for 1.5\,h with 512 robots on \sysname, drives the shortest path, 1040\,m in 655\,s, 9\,\% shorter than the released checkpoint. At the same 2\,h budget, the SAC policy trained on \sysname drives 22\,\% less than the one trained with ARiADNE's code, and it matches the released checkpoint with a fraction of that checkpoint's training time. \sysname unlocks the use of on-policy algorithms like PPO by reducing the time taken for data collection.
We also deploy the same PPO policy in the indoor corridors world of the
same simulator \citep{cao2022autonomous}. It is a 130$\times$100\,m map of long
corridors and lobby areas
(Fig.~\ref{fig:indoor}). The conclusion from
\texttt{forest} therefore carries over to a structurally different
environment: our policy matches the baseline's path length at a fraction of the training budget.

\begin{figure*}[t]
    \centering
    \includegraphics[page=1, width=0.8\textwidth, trim=1pt 240pt 1pt 1pt, clip]{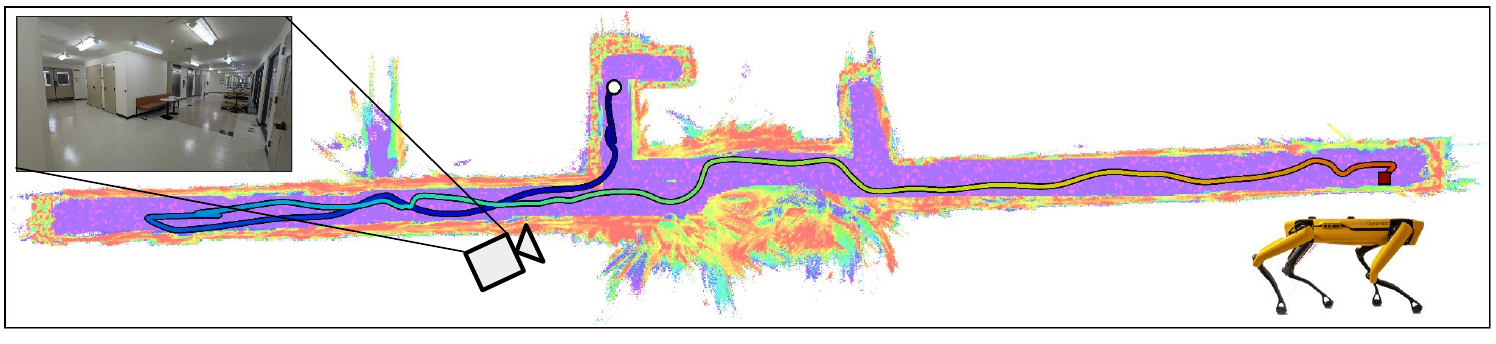}
    \caption{\small\textbf{The simulation-trained PPO policy successfully explores a real indoor map on the Spot robot using on-board cameras only.} The top-down point cloud (colour = height) with the $157\,\mathrm{m}$ trajectory is shown. Insets: an image of the corridor and the robot.}
    \label{fig:real_exploration}
\end{figure*}

Finally, we deploy the same PPO policy on the Spot robot exploring an office floor (Fig.~\ref{fig:real_exploration}). The policy was trained entirely on 2-D grid-world maps and has never seen real data, yet it transfers directly, because its observation is the graph and not the raw sensor feed. The robot's path length was 157\,m. In certain areas, the cameras saw through glass doors and free space was mapped inside closed rooms and nodes were sampled there. The policy initially targeted these nodes for their frontier utility, but once the local planner could not reach them, the graph builder's connectivity check dropped them from the robot's graph component and the policy moved on to the reachable frontiers.

\section{Conclusion}
\label{sec:conclusion}
We presented \sysname: an open-source, GPU-resident navigation-graph library that serves as one design tool for robot deployment, navigation and policy learning. It maintains frontier detection, node sampling, incremental merging, edge validation and attribute layers on the GPU. The same GPU kernels build a 3-D graph, host frontier and semantic layers that classical planners read directly, correct a drifted graph after loop closure in milliseconds, and step a thousand simulated robots per tick. Using \sysname, training RL-based exploration policies reaches the quality of the baseline graph-based policies in an order of magnitude less wall-clock time. The policy outperforms classical planners on held-out maps, TARE in simulation, and SoTA methods like ARiADNE and GRATE at similar training budgets. It also transfers to a ground vehicle in a high-fidelity simulator and in the real world. Future work will replace the per-platform tuning of sampling parameters with learned sampling of the nodes in the graph and handle dynamic obstacles. As node attributes are tensors, the same batched
environment can potentially extend to image-conditioned policies, with a vision encoder
writing per-node embeddings from an RGB feed like our semantic layer already does
(Fig.~\ref{fig:layers}B).

\section{Acknowledgments}
\label{sec:ack}
The authors used Claude/Codex to assist with
code prototyping and debugging, manuscript revision, and formatting of Figs. The related works content used AI verification. The authors take full responsibility for the final content.
{\footnotesize
\setlength{\bibsep}{0pt plus 0.2pt}
\bibliographystyle{IEEEtran}
\bibliography{references}
}

\end{document}